\documentclass[10pt,twocolumn,letterpaper]{article}

\usepackage{wacv}
\usepackage{times}
\usepackage{epsfig}
\usepackage{graphicx}
\usepackage{amsmath}
\usepackage{amssymb}
\usepackage{multirow}
\usepackage{multicol}
\usepackage{color,colortbl}
\usepackage[table]{xcolor}
\usepackage[accsupp]{axessibility}  

\newcommand{\mm}[1]{\textcolor[rgb]{0,0,0}{#1}}    
\newcommand{\white}[1]{\textcolor[rgb]{1,1,1}{#1}}    

\def\wacvPaperID{21} 

\wacvfinalcopy 

\ifwacvfinal
\def\assignedStartPage{9876} 
\fi

\ifwacvfinal
\usepackage[breaklinks=true,bookmarks=false]{hyperref}
\else
\usepackage[pagebackref=true,breaklinks=true,colorlinks,bookmarks=false]{hyperref}
\fi

\ifwacvfinal
\else
\fi

\begin{document}

\title{\vspace{-30pt} Small or Far Away? Exploiting Deep Super-Resolution and Altitude Data for Aerial Animal Surveillance\vspace{-10pt}}


\author{\white{.......} Mowen Xue 
\white{.............} Theo Greenslade 
\white{.......} Majid Mirmehdi 
\white{..........} Tilo Burghardt\\
{\white{...}{\tt\footnotesize dt20957@bristol.ac.uk}  \white{.} {\tt\footnotesize tg17437@bristol.ac.uk} \white{.}{\tt\footnotesize majid@cs.bris.ac.uk} \white{..} {\tt\footnotesize tilo@cs.bris.ac.uk}} \footnotesize\ \\
Dept of Computer Science, University of Bristol, Bristol, BS8 1UB, UK
}

\maketitle

\thispagestyle{empty}

\begin{abstract}
   Visuals captured by high-flying aerial drones are increasingly used to assess biodiversity and animal population dynamics around the globe. Yet, challenging acquisition scenarios and tiny animal depictions in airborne imagery, despite ultra-high resolution cameras, have so far been limiting factors for applying computer vision detectors successfully with high confidence. In this paper, we address the problem for the first time by combining deep object detectors with super-resolution techniques and altitude data. In particular, we show that the integration of a holistic attention network based super-resolution approach and a custom-built altitude data exploitation network into standard recognition pipelines can considerably increase the detection efficacy in real-world settings. We evaluate the system on two public, large aerial-capture animal datasets, SAVMAP and AED. We find that the proposed approach can consistently improve over ablated baselines and the state-of-the-art performance for both datasets. In addition, we provide a systematic analysis of the relationship between animal resolution and detection performance. We conclude that super-resolution and altitude knowledge exploitation techniques can significantly increase benchmarks across settings and, thus, should be used routinely when detecting minutely resolved animals in aerial imagery.\vspace{-18pt}
\end{abstract}


\begin{figure}[t]
\begin{center}
\vspace{-10pt}
\includegraphics[width=237pt,height=360pt]{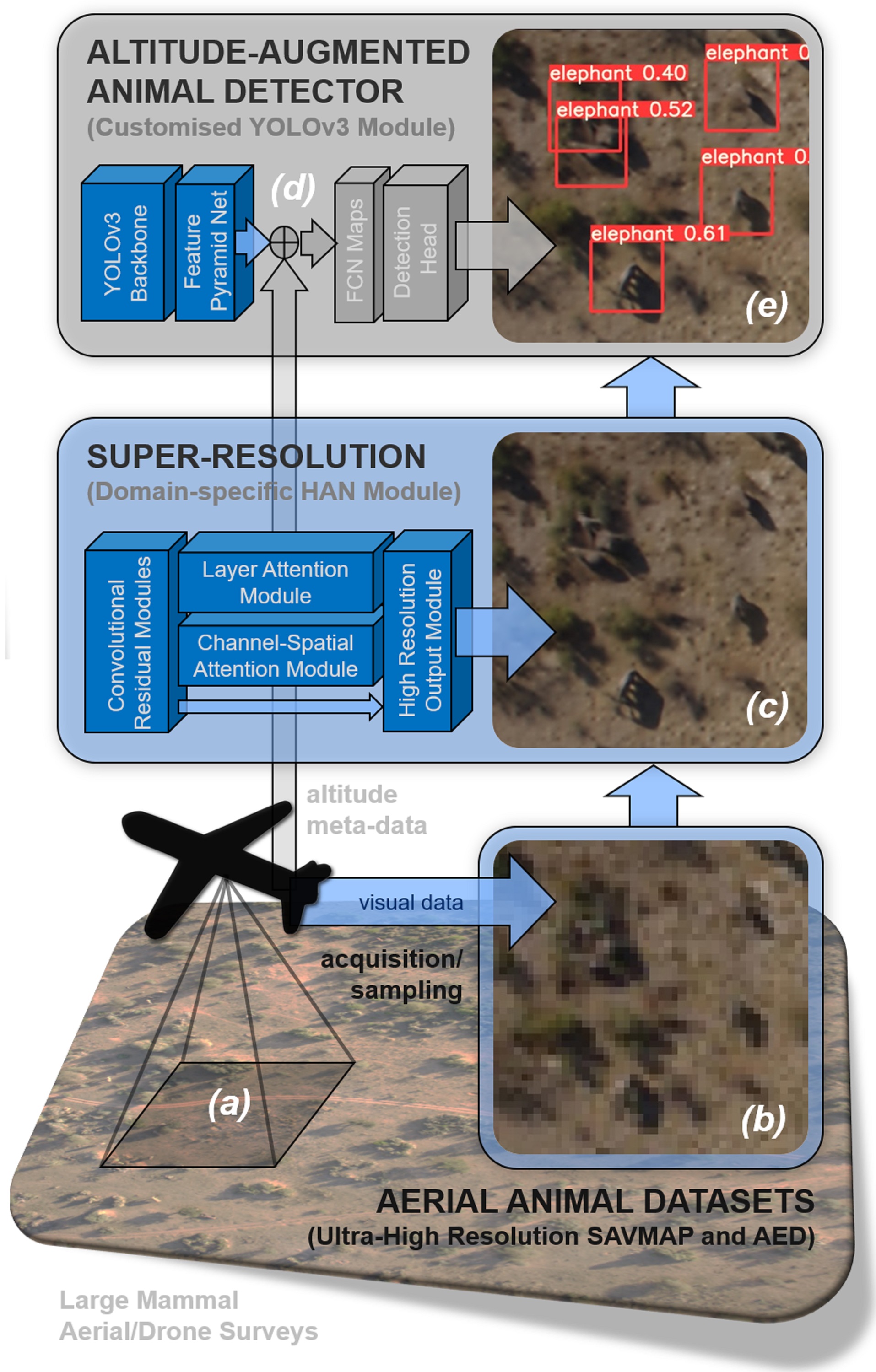}\vspace{-8pt}
\end{center}
   \caption{\textbf{Conceptual Overview.} Our approach integrates super-resolution and altitude data exploitation directly into deep animal detection pipelines for aerial survey applications. \textbf{\textit{(a)}} During aerial image capture both ultra-high resolution RGB stills (blue) and associated altitude data (grey) is recorded. \textbf{\textit{(b)}} We experiment with resulting original and systematically downsampled visuals. \textbf{\textit{(c)}} A domain-trained holistic attention network is used to super-resolve the imagery,  enhancing minutely resolved animal depictions. \textbf{\textit{(d)}}~Altitude data is then used as additional network input to effectively constrain the valid animal scale and appearance. \textbf{\textit{(e)}}~A custom-trained baseline network (YOLOv3) finally performs animal detection on the altitude-aware super-resolved inputs. For the public SAVMAP and AED datasets the setup proves highly effective, improving benchmarks beyond baselines and prior works.}\vspace{-25pt}
\label{fig:conceptual overview}
\end{figure}

\section{Introduction}
\textbf{Motivation and Aerial Surveys.} Collecting regular wildlife census information through timely and accurate population surveillance~\cite{Smyser_2016_aerial_line_transect_surveys,Hodgson_2018_drones_count_accurately, Rylance_2017_contribution_of_tourism} is crucial in understanding how animal populations move and change~\cite{Linchant_2015_uas_the_future}, and how conservation efforts can be conducted to counter-act environmental degradation~\cite{Skogen_2018_climate_change_concern} and species decline~\cite{Bouche_2011_will_elephants_disappear,Ceballos_2015_sixth_mass_extinction}. Whilst surveillance on human populations opens a multitude of ethical concerns, surveillance for the purpose of animal protection is an ethical imperative. Manual surveys~\cite{Shiu_2014_acoustic_monitoring,Schlossberg_2016_testing_aerial_surveys} are often expensive though, have limited site access~\cite{Wang_2019_review_UAS_sat_aircraft}, and may even expose staff to poacher threats~\cite{OGrady_2020_protecting_rhinos} or transport risks~\cite{Linchant_2015_uas_the_future}. 
Recently, the use of unmanned aerial vehicles~(UAVs) with ultra-high resolution cameras has emerged as a cost-effective alternative for various survey types~\cite{Wang_2019_review_UAS_sat_aircraft,Hodgson_2018_drones_count_accurately,Vermeulen_2013_aerial_survey_of_elephants}. However, the detection of scarce and often minutely resolved animals in vast amounts of highly variable environmental image content (see Fig.~\ref{fig:conceptual overview}\textit{(a)}) still poses a significant challenge.

\textbf{Animal Detection in Aerial Imagery.} Recent deep learning approaches that perform object detection for animal recognition in aerial imagery have been applied with some success~\cite{Rey_2017_detecting_with_uavs_and_crowds,Kellenberger_2018_detecting_mammals,Duporge_2020_satellite_elephants}, but in contrast to other fields of vision, like image classification~\cite{classification_benchmark}, autonomous driving~\cite{autodriving_benchmark}, benchmarks lack well behind human performance. In fact, even tiny object detection in less cluttered and variable environments poses an ongoing challenge for current object detection methods~\cite{Yu_2020_tiny_object_challenge}. Whilst topics like dataset imbalance and animal scarcity~\cite{Kellenberger_2018_detecting_mammals}, domain variability and transfer~\cite{Wang_2019_review_UAS_sat_aircraft, Kellenberger_2020}, and semi-supervision and active learning~\cite{Rey_2017_detecting_with_uavs_and_crowds,Kellenberger_2019_few_clicks,Kellenberger_2019_half_a_percent} have well been studied in the domain of aerial animal detection, the problem of minute animal resolution~\cite{Wang_2019_review_UAS_sat_aircraft} has not been tackled explicitly to date. As stated, animal appearance information often resides in only a few dozen pixels per animal for the majority of aerial datasets. Similar to recent works in other domains~\cite{appliedSR1, appliedSR2, appliedSR3}, we observed that domain-trained super-resolution~(SR) techniques can recover valid animal-specific appearance information in many cases. In addition, we noted that virtually all aerial datasets are tagged with altitude information which could be used as an inference-basis to relate and constrain expected animal sizes in the images. 

\textbf{Paper Concept.} Bringing these two ideas together, we propose directly combining deep object detectors with super-resolution techniques and altitude data in a single recognition pipeline as shown in Fig.~\ref{fig:conceptual overview}. In particular, we will show that the integration of a holistic attention network (HAN) super-resolution approach and a altitude data exploitation network integrated into a \mm{baseline detector} pipeline can significantly increase the detection efficacy of tiny animal detection in aerial imagery. 

 \textbf{Main Contributions.} Our key contributions can be summarised as follows: 1) We introduce a new animal detection approach for aerial surveys that integrates HAN-based super-resolution and altitude data into a \mm{baseline detector} pipeline. 2) We evaluate our method on the two main, large aerial-capture animal datasets SAVMAP and AED, outperforming baselines and the state-of-the-art. 3) We perform detailed ablations and provide a systematic analysis of the relationship between animal resolution and detection performance in the system.

\section{Background}

\subsection{Deep Learning for Animal Detection}

 \textbf{Deep Object Detectors.} Deep learning for object detection forms an active and extensive research field in computer vision. Many detector designs have been proposed in the recent past~\cite{EfficientDet, Context_R-CNN, end-to-end_semi_supervised} with most common approaches following either a one-stage~\cite{Redmon_2018_YOLOv3,Lin_2018_RetinaNet,Liu_2016_SSD} or two-stage network design~\cite{Girshick_2014_RCNN,He_2014_sppnet,Girshick_2015_fast_rcnn,Ren_2016_faster_rcnn}. Applications to animal biometrics~\cite{kuhl2013} most often encompass species detection in camera trap imagery~\cite{Schneider_2018_DL_camera_trap_data, Norouzzadeh_2017_camera_traps_with_DL, Willi_2018_camera_traps,camtrap2019} or manual photography~\cite{brust2017,manually_photograph}, as well as in captive settings~\cite{pig_detection} and agricultural inspection~\cite{cattle_detection}.

\begin{figure}[t]
\begin{center}
\includegraphics[width=237pt,height=180pt]{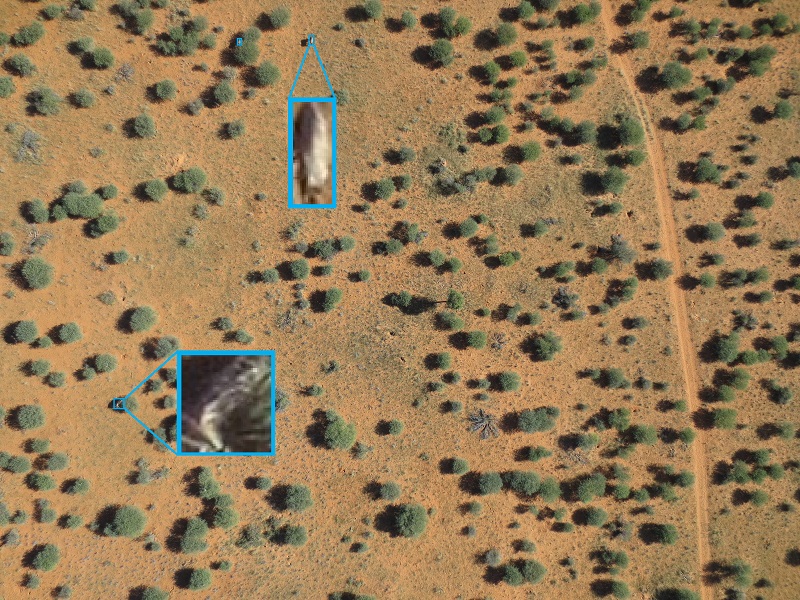}
\includegraphics[width=237pt,height=160pt]{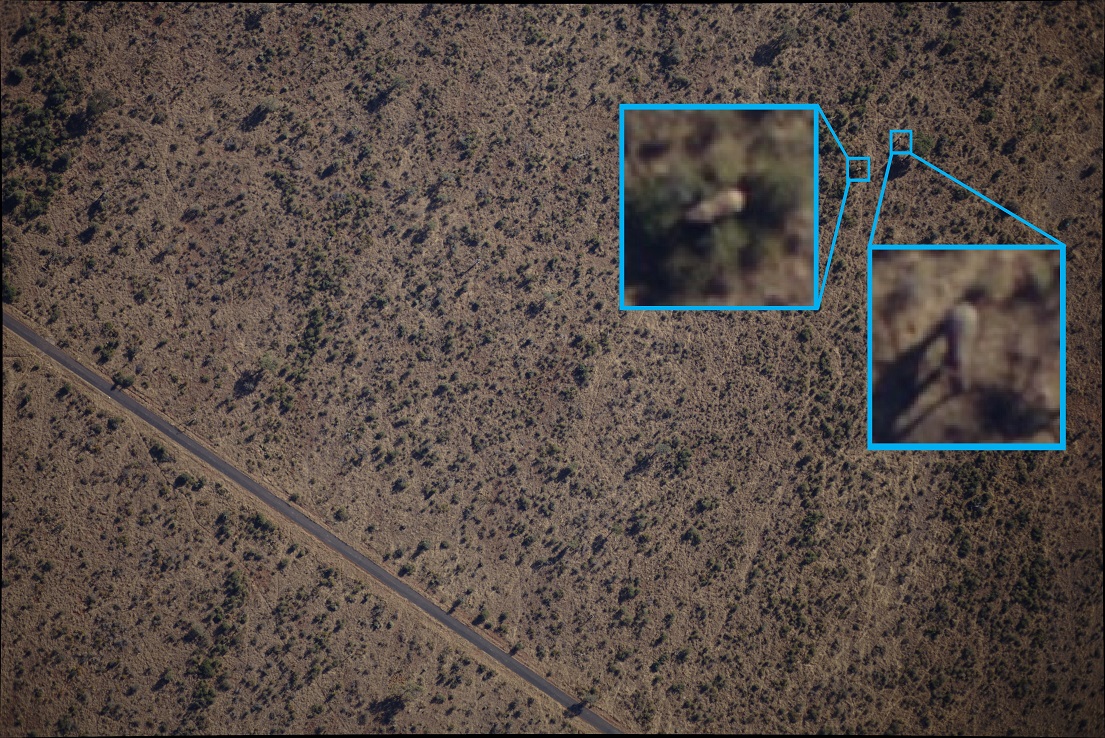}\vspace{-8pt}
\end{center}
   \caption{\textbf{Challenging Aerial Animal Imagery.} Representative images in original resolution selected \textbf{\textit{(top)}} from the 654 ultra-high resolution SAVMAP images produced by high-flying drones, and \textbf{\textit{(bottom)}} from the 2,074 images of the AED elephant dataset. Ground truth animal annotations are highlighted and some zoom-in is provided for better appreciation of the visuals. Note the vastness of the depicted environment and the low and challenging animal resolutions associated to the data.}\vspace{-5pt}
\label{fig:datasets}
\end{figure}

 \textbf{Aerial Animal Detection.} Research into applying object detection to aerial imagery of animals is still rare, but recent works have picked up pace with a focus on species such as whales~\cite{guirado2019whale},  cattle~\cite{Barbedo_2019_Cattle_in_UAV,cattle2019}, and other large mammals~\cite{Rey_2017_detecting_with_uavs_and_crowds,Kellenberger_2018_detecting_mammals}. Utilising the  SAVMAP aerial dataset~\cite{Reinhard_2015_SAVMAP} containing large African mammals in vast savannah environments~(see Fig.~\ref{fig:datasets}\textit{(top)}), Kellenberger et al.~\cite{Kellenberger_2018_detecting_mammals} showed recently that deep learning approaches outperform traditional vision techniques~\cite{Rey_2017_detecting_with_uavs_and_crowds} in this domain, too. Their system used a ResNet-18 backbone with two multi-layer {perceptrons} (MLPs) added with nonlinear activations (ReLU), dropout {regularization}, and Softmax activation. 
 To further improve the system, curriculum learning, hard negative mining, a border class, and a new Census-oriented evaluation protocol were introduced by the group. They showed that 1) this model could still be effective when trained using weakly-supervised learning with only a small number of fully annotated labels~\cite{Kellenberger_2019_few_clicks}, and 2) how this model could be transferred to new datasets using active learning~\cite{Kellenberger_2019_half_a_percent}. 
 
 Apart from SAVMAP, the Aerial Elephant Dataset~(AED)~\cite{Naude_2019_AED} has recently been released as another public large-scale aerial drone dataset for animal detection, albeit providing slightly higher animal resolutions~(see Fig.~\ref{fig:datasets}\textit{(bottom)}). To benchmark this dataset, MobileNet~\cite{mobilenets} was adapted to a fully convolutional architecture by the authors~\cite{Naude_2019_AED} to perform image segmentation and ultimately elephant detection. Most recently, Duporge et al.~\cite{Duporge_2020_satellite_elephants} applied deep object detectors to find elephants in non-public, high-resolution satellite data (copyrighted by Maxar Technologies and the European Space Agency) to compare manual detection benchmarks against a standard two-stage Faster Region Convolutional Neural Network (F-RCNN) model. Due to the limitations of accessing satellite datasets, this paper will focus evaluation efforts on the publicly available drone datasets SAVMAP and AED to aid transparent scientific comparability and universal reproducibility. To the best of our knowledge, none of the published animal detection approaches for aerial data has so far utilised altitude data or addressed very low animal resolution explicitly.

 \textbf{Tiny Object Detection in Aerial Images.} Current research into tiny object detection~\cite{Yu_2020_tiny_object_challenge} has resulted in many frameworks to address this 'few pixel detection challenge'. Most aerial datasets used to evaluate approaches contain vehicles, such as cars or planes, captured via satellites~\cite{Pang_2019_r2cnn, Shermeyer_2019_SR_satellite}. 
 Approaches that address the resolution challenge include the use of feature pyramid networks~\cite{Lin_2017_feature_pyramid_network}, hard mining methods~\cite{Shrivastava_2016_hard_mining} and more recently, attention base mechanisms~\cite{Lim_2021_tiny_attention}. Benchmarking network architectures for tiny object detection, a very recent review~\cite{review2021} concluded that F-RCNN and YOLOv3 ('You-Only-Look-Once')~\cite{Redmon_2016_YOLO,Redmon_2018_YOLOv3} currently perform strongest for the task across most performance statistics. Torney et al.~\cite{Torney_2019_comparision_DL_citizens} even showed that YOLOv3 could achieve comparable results with human annotators when detecting wildebeest in images from the Zooniverse-driven~\cite{zooniverse} Serengeti Wildebeest Count using aerial data from low-flying drones~\cite{low2020}. Although similar in content to usual survey data like SAVMAP, these images are taken from an up to five times lower altitude and show animals resolved significantly larger than for most common surveys. Nevertheless, following these leads~\cite{review2021,Torney_2019_comparision_DL_citizens} we base our core architecture around a YOLOv3 backbone to leverage its noted applicability to wildlife detection and its proven performance on detecting tiny object content.  
 
\subsection{Deep Learning for Super-Resolution}
Only recently has tiny object detection in aerial images been addressed via the application of super-resolution~\cite{Courtrai_2020_SR_tiny_detection}. This approach builds on a long-standing research thread in performing super-resolution via deep learning initiated by works such as SRCNN~\cite{Dong_2014_SRCNN}. Many architectures followed this ground-breaking work using more and more modern deep learning techniques to improve these results. The recursive convolutional networks DRCN~\cite{Kim_2016_DRCN} and DRRN~\cite{Tai_2017_DRRN} were introduced, a pyramidal framework was used in the LapSR network~\cite{Lai_2017_lapSR}, and a generative adversarial network (GAN) was used in work on SRGAN~\cite{Ledig_2017_srgan}. Generally, residual learning and particularly deep architectures with large receptive fields produce particularly strong super-resolution performance~\cite{Niu_2020_HAN}.

Attention mechanisms have further improved benchmarks in recent years~\cite{Zhang_2018_RCAN, Woo_2018_CASA,Kim_2018_RAM,Hu_2019_channel_spatial_modulation}. Thus, in this research we utilise the current HAN super-resolution approach~\cite{Niu_2020_HAN}, which indeed incorporates two attention components -- a layer attention module (LAM) and a channel-spatial attention module (CSAM). It performs state-of-the-art single image super resolution (SISR). Again, to the best of our knowledge, super-resolution has never been investigated for detecting animals in aerial imagery.

\section{Datasets}

This work is evaluated on two of the biggest public aerial drone datasets containing animals, that is the SAVMAP dataset~\cite{Reinhard_2015_SAVMAP} and the AED dataset~\cite{Naude_2019_AED}.  Both represent real-world conditions that realistically reflect the challenges of performing aerial surveys. They cover African landscapes with manually annotated labels for the location of wildlife.  

\textbf{SAVMAP:} The SAVMAP dataset was taken in the Kuzikus Wildlife Reserve between May 12 and May 15 2014. Kuzikus is a private wildlife park covering an area of $103 km^{2}$ located in eastern Namibia. There are more than 20 species and 3,000 large mammals in this park, including Common Elands (Taurotragus oryx), Greater Kudus (Tragelaphus strepsiceros), Gemsboks (Oryx gazella), Hartebeests (Alcelaphus buselaphus), Gnus (Connochaetes gnou and Connochaetes taurinus) and others~\cite{Ofli_2016_combining_human_computing_and_ML}. The dataset covers five flights using an ultra-high resolution Canon Powershot camera fixed on the aircraft. It contains 654 images, resolved at $3000\times4000$~pixels. Following Kellenberger et al \cite{Kellenberger_2018_detecting_mammals}, we used their training-validation-test split of $70\%-10\%-20\%$. Original MicroMapper crowd sourced labels~\cite{Ofli_2016_combining_human_computing_and_ML} were error-corrected and improved by Kellenberger et al.~\cite{Kellenberger_2018_detecting_mammals}. The detection ground truth is formed by 1,183 tight animal bounding boxes with an average size of $25\times23$ pixels~(see Fig.~\ref{fig:datasets}\textit{(top)}). 

\textbf{AED:} This dataset contains aerial images of African elephants (Loxodonta Africana) captured between 2014 and 2018. Resolving animals at slightly higher resolutions compared to SAVMAP, the dataset has 2,074 images containing 15,581 elephants~\cite{Naude_2019_AED} with a train-test split of $80\%-20\%$. Canon 6D digital single-lens reflex cameras were attached to a SkyReach BushCat light-sport aircraft to produce the dataset. To maximise the image size, imagery was acquired using three cameras: one pointing straight down, and two pointing out left and right by 20 degrees each, all controlled by a Raspberry Pi to synchronise the capture. The images were acquired in 5 different wildlife reserves in Africa in 8 separate campaigns: Hluhluwe-iMfolozi Park and Phinda Private Game Reserve in central KwaZulu-Natal, South Africa, the Northern Tuli Game Reserve in the Tuli Block, Botswana, NG26 concession in the Okavango Delta, Botswana, Bwabwata and Mudumu national parks in the Zambezi strip, Namibia and the Madikwe game reserve in the North-West province, South Africa. The dataset contains images captured at various times of day from sunrise to sunset and in both the wet season and dry season. The labels provided in the dataset are coordinates of the centre of the elephants in the images. For comparability with a bounding box paradigm, we also defined approximate bounding boxes of $100\times100$ pixels around the centre coordinates even though some elephants (e.g. juveniles) in the dataset take up a smaller area~(see Fig.~\ref{fig:datasets}\textit{(bottom)}).

\section{Method}

Fig.~\ref{fig:conceptual overview} outlines the proposed  recognition pipeline: acquired ultra-high resolution RGB input is first super-resolved via the HAN Resolution Enhancement Module and then fed forward into an Altitude-augmented Module which performs animal detection.   

\begin{figure}[t]
\begin{center}
\includegraphics[width=237pt,height=225pt]{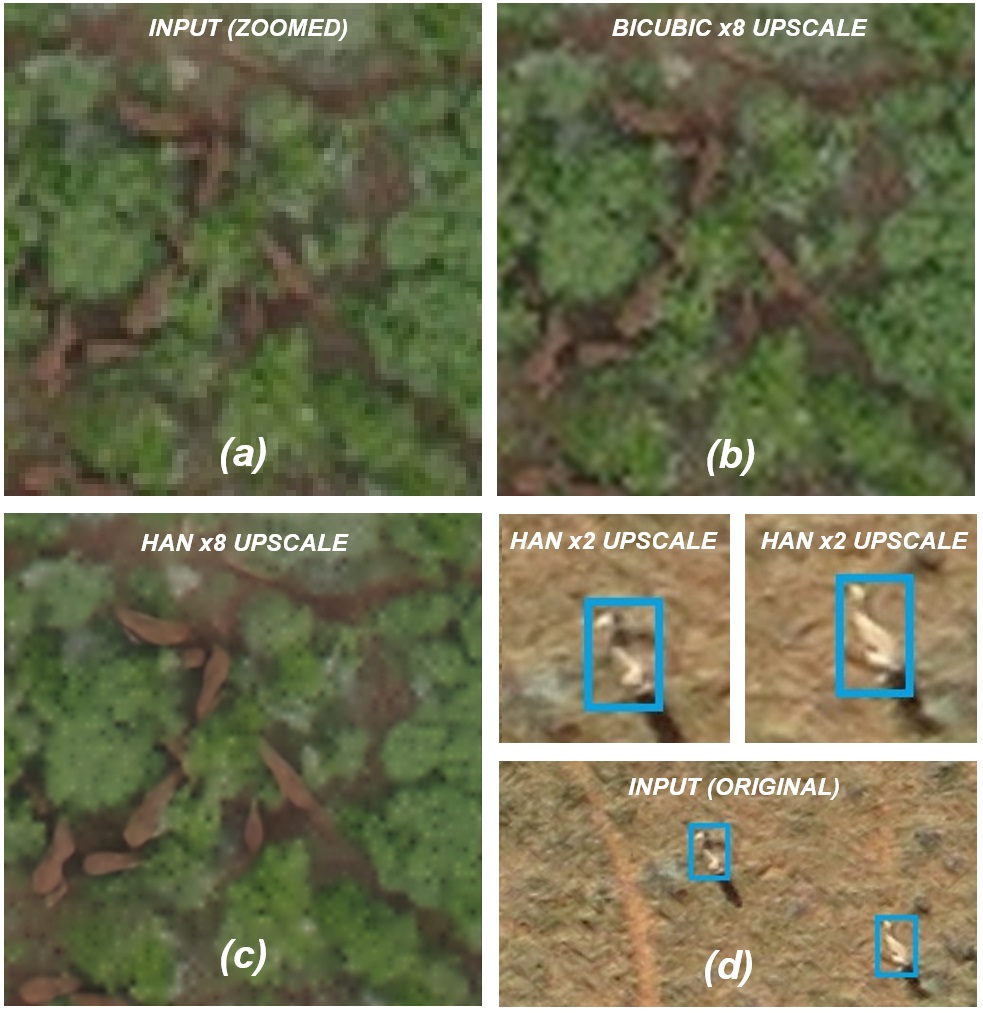}\vspace{-8pt}
\end{center}
   \caption{\textbf{HAN Super-Resolution of Aerial Animal Content.} Qualitative example depictions of super-resolution results applied to aerial animal imagery. \textbf{\textit{(a)}} First, we show an AED data patch sampled at low animal resolution upscaled via \textbf{\textit{(b)}} an 8-fold bicubic interpolation baseline, which is visually outperformed by \textbf{\textit{(c)}}~our domain-trained HAN super-resolution module producing a significantly clearer image at 8-fold upscale. \textbf{\textit{(d)}}~Secondly, we show a 2-fold upscaling application of our super-resolution component to a SAVMAP image patch with ground truth annotations superimposed. Note the enhancement of animal recognisability across examples. Our quantitative results in Tables~\ref{table:results} and~\ref{table:analysis} show that these qualitative observations regarding super-resolution are indeed aligned with improved detection performance.}\vspace{-5pt}
\label{fig:superresolution}
\end{figure}

\subsection{HAN Resolution Enhancement Module}
\label{sec:sr}
Inspired by RCAN~\cite{Zhang_2018_RCAN}, {we use HAN resolution enhancement~(see Fig.~\ref{fig:conceptual overview}\textit{(c)}) exactly as described in~\cite{Niu_2020_HAN} with its four fundamental parts:} 1)~a feature extraction backbone, followed by 2)~LAM and 3)~CSAM holistic feature weighting and, finally, 4)~the image reconstruction block generating super-resolved content. To perform feature extraction, first a convolutional layer extracts shallow features from the low-resolution input, which are subsequently passed through a backbone made up of several residual groups to form content appearance features. Then, the  LAM examines correlations between layers to emphasise hierarchical features in the image adaptively. These feature correlations are often ignored by other current SISR methods that use CNNs which often results in texture details in output images being smoothed, which is avoided here. To incorporate channel-spatial attention dependencies, we also use CSAM to selectively capture more informative features by learning across all channels. The super-resolved output finally produced by the reconstruction block~(see Fig.~\ref{fig:superresolution} for examples) serves as input to a subsequent detection module.

\subsection{Altitude-augmented Module}
\label{sec:altitude}
\mm{Our proposed altitude-augmented module explicitly incorporates} altitude information by feature concatenation leading into the detection head. Incoming super-resolved content is processed via the DarkNet~\cite{Redmon_2018_YOLOv3} backbone and a subsequent feature pyramid network. \mm{The latter provides content analysis at different scale levels -  our key ingredient to enhancing tiny object detection. The resulting feature map is flattened into a long vector. We concatenate the altitude data, for example altitude $\mathcal{A}=1496.68$ (in metres) represented as a 32bit float scalar,  at the end of the feature vector to form the final feature representation, which is mapped via two fully connected layers before being fed into the detection head.} The main components of the altitude-augmented module is outlined in~Fig.~\ref{fig:conceptual overview}\textit{(d/e)}).  The code  for this work is available at \url{https://github.com/Mowen111/SALT}.


\section{Implementation and Experimental Setup}
\label{sec:imp}
\textbf{Domain-specific Super-Resolution Training.} We first downsampled the training image portion of the AED dataset -- which as discussed is resolved higher than SAVMAP -- by factors 2, 4 or 8, and then use these truly low-resolution images and corresponding higher-resolution images to train the super-resolution module. The initial learning rate was set to 1e-4. We used step decay with a learning rate decay factor of 0.5. The optimiser used for training was Adam~\cite{adam} where beta1 was set to 0.9, beta2 was 0.999, and epsilon was 1e-8. We use weight decay with a value of 0.0001. We train the system for 50 epochs in total which resulted in a PSNR score on the validation dataset of 37.31, 31.68 and 27.30 for networks that scale by a factor of 2, 4 and 8 respectively.

\begin{figure}[b]
\begin{center}
\includegraphics[width=240pt,height=130pt]{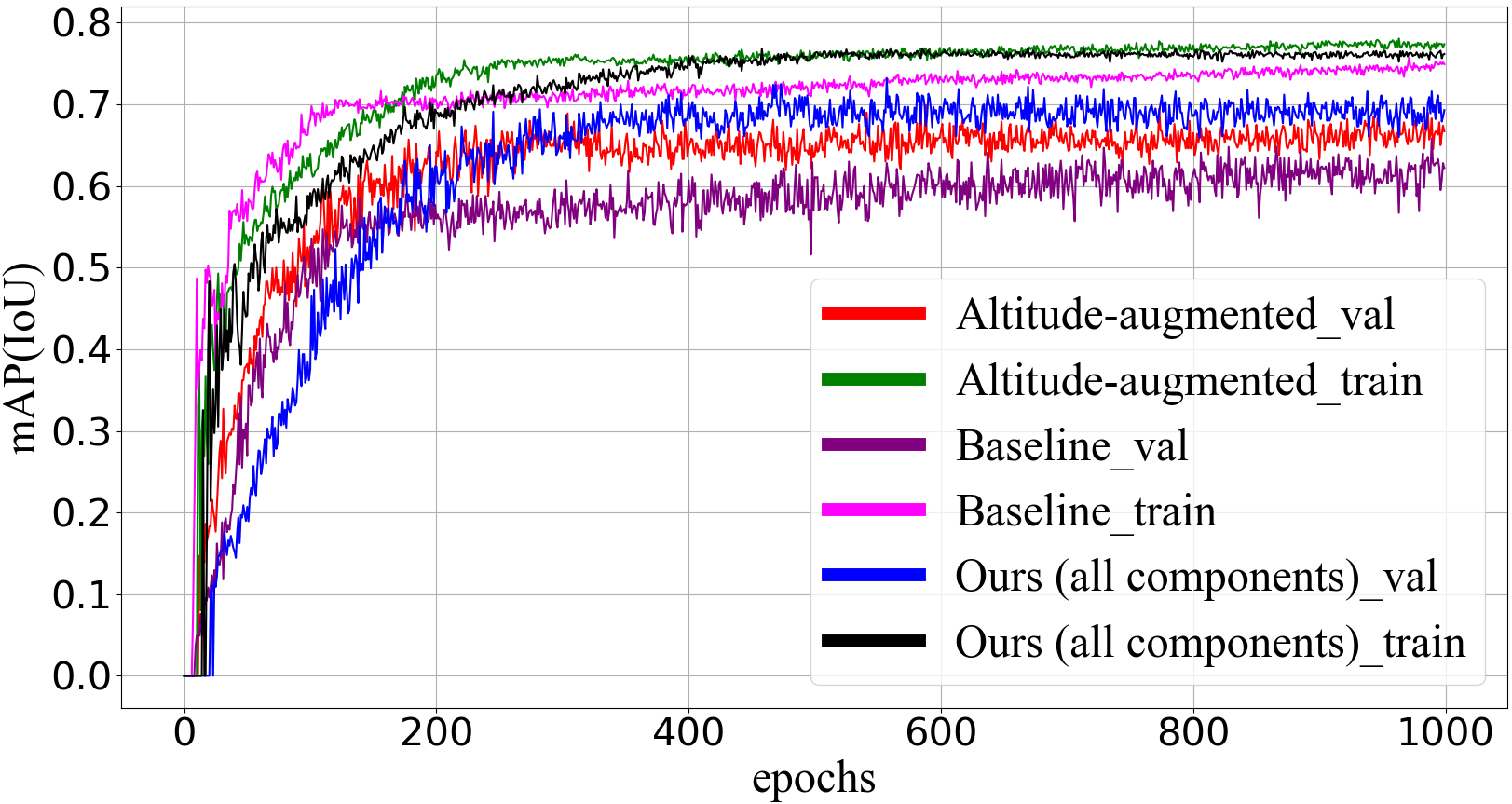}\vspace{-8pt}
\end{center}
   \caption{\textbf{Training Evolution.}  Depicted is the development of SAVMAP training and validation mAP(IoU) results over 1,000 epochs of the SGD optimisation process for \mm{our baseline (YOLOv3~\cite{Redmon_2018_YOLOv3})} \textbf{\textit{(purple and magenta)}}, the altitude-augmented model \textbf{\textit{(red and green)}}, and our proposed model \textbf{\textit{(blue and black)}}.}\vspace{-5pt}
\label{fig:train}
\end{figure}

\textbf{Altitude-augmented Training.} With a domain-trained super-resolution module in hand, we apply the learned resolution upscaling to both the SAVMAP and AED datasets. Subsequently, per dataset we train the altitude-augmented module on the resolution-enhanced visual input (sub-patched at a resolution of $512\times 512$ pixels) and associated altitude meta-data using an initial learning rate of 1e-4, weight decay of 0.001, and momentum of 0.9. Standard Stochastic Gradient Descent~(SGD)~\cite{SGD} optimisation is used to train the network for 1,000 epochs~(see~Fig.~\ref{fig:train}).

\begin{figure}[t]
\begin{center}
\includegraphics[width=240pt,height=130pt]{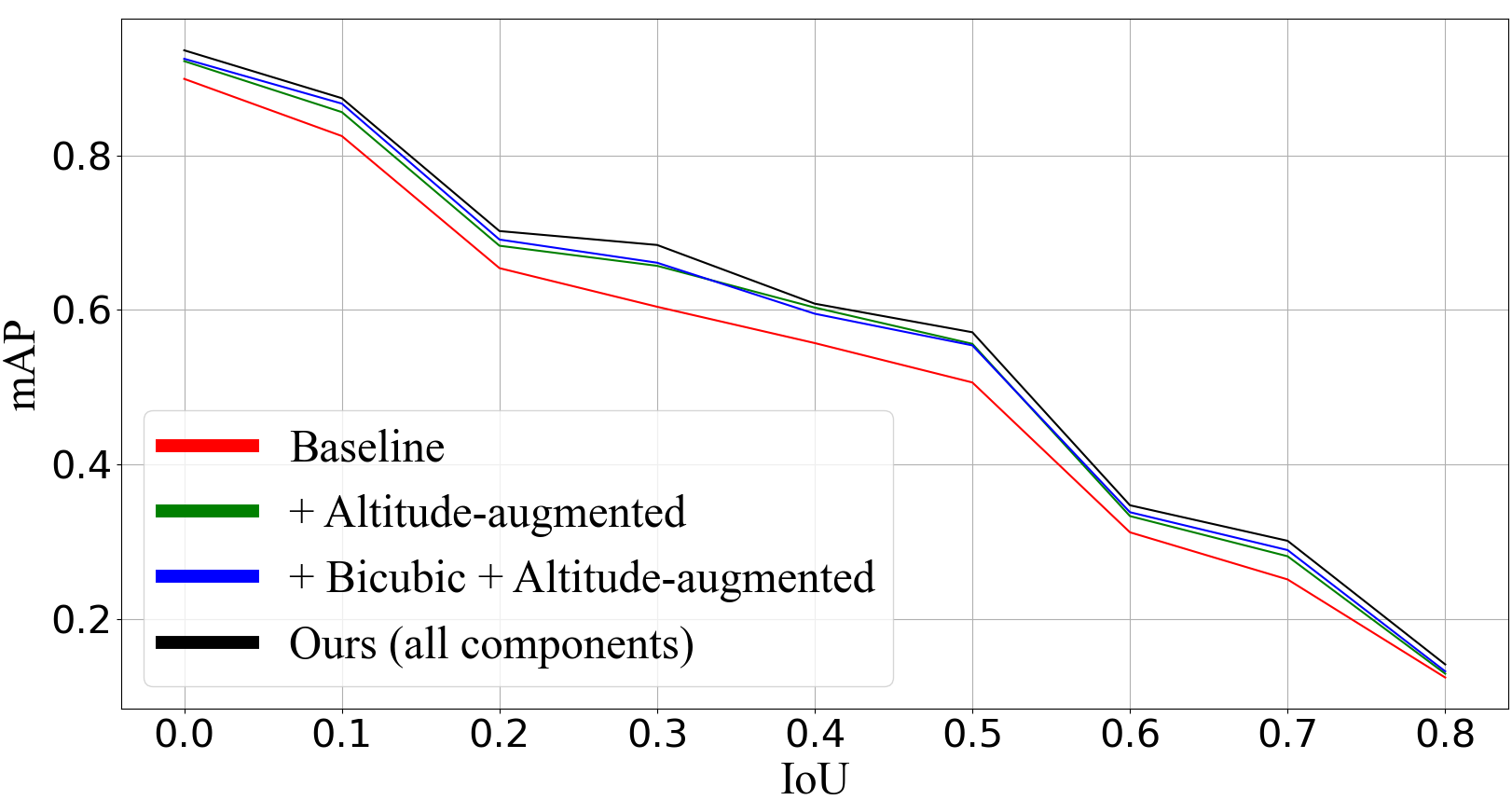}\vspace{-8pt}
\end{center}
   \caption{\textbf{Result Consistency.} \mm{Expectedly, mAP on SAVMAP test data declines as the IoU threshold for detection increases. However, our proposed method consistently performs better than other methods across the entire spectrum.}}\vspace{-5pt}
\label{fig:IoU_mAP}
\end{figure}

\textbf{Inference Pipeline.} The testing portion of each of the datasets are used and divided into patches (at a resolution of $512\times 512$ pixels). For each patch, super-resolution is applied and before feeding the enhanced visual together with altitude meta-data into the altitude-augmented module to yield animal detections. Following~\cite{Redmon_2018_YOLOv3}, for visualisations and tests we use a detection confidence threshold of $0.1$ across all compared setups.

\begin{table*}[ht!]
\begin{center}
\begin{tabular}{|c|c|l|c|c|c|}
\hline
\textbf{Dataset} &   \textbf{Row} & {\textbf{Method}}   &    \textbf{Operational Resolution}  & \textbf{mAP}(IoU) & \textbf{mAP}(Che)      \\ \hline\hline
\multirow{5}{*}{\small{SAVMAP \cite{Reinhard_2015_SAVMAP}}} &  1 &  Kellenberger et at.\cite{Kellenberger_2019_few_clicks} &  \small{\multirow{3}{*}{$512\times 512$}}                          & 0.588                & —                  \\ \cline{2-3} \cline{5-6}
                         & 2 & Baseline     &       &                            0.654                & 0.855                \\ \cline{2-3} \cline{5-6}
                         & 3 & + Altitude-augmented    &           & 0.683                & 0.875                \\ \cline{2-6} 
                         & 4 & + Bicubic + Altitude-augmented & \small{\multirow{2}{*}{$512\times 512 \rightarrow 1024\times 1024$} }                          & 0.691                & 0.886                \\ \cline{2-3} \cline{5-6}
                         & 5 & \textbf{Ours (all components)}  &                   & \textbf{0.702} &\textbf{0.892} \\ 
                         \hline\hline
\multirow{5}{*}{\small{AED \cite{Naude_2019_AED}}}    &  6 & Naude et at.\cite{Naude_2019_AED}        &  \small{\multirow{3}{*}{$512\times 512$}}                               & —                  & 0.890                 \\ \cline{2-3} \cline{5-6} 
                         & 7 & Baseline   &      &          0.721                & 0.915                \\ \cline{2-3} \cline{5-6} 
                         & 8 & + Altitude-augmented &                  & 0.755       & 0.934                \\ \cline{2-6} 
                         & 9 & + Bicubic + Altitude-augmented & \small{\multirow{2}{*}{$512\times 512 \rightarrow 1024\times 1024$}}                             & 0.763                & 0.946                \\ \cline{2-3} \cline{5-6} 
                             &  10 & \textbf{Ours (all components)}  &                         &  \textbf{0.778} & \textbf{0.955} \\  
                         \hline
\end{tabular}
\end{center}
\caption{\textbf{Result Overview.} We compare mAP results (showing both IoU and Chebychev calculations as discussed in~Sec.~\ref{sec:imp}) on the testing portion of each of the two datasets SAVMAP and AED. Previously published state-of-the-art benchmarks are given in rows 1 and 6. \mm{We use the state-of-the-art standard YOLOv3~\cite{Redmon_2018_YOLOv3} as our baseline~(see rows 2 and 7).} Augmenting baseline with altitude meta-data~(see~Sec.~\ref{sec:altitude}) can further improve animal detections~(rows 3 and 8). Scaling image resolutions up using bicubic interpolation before detection consistently improves benchmarks again~(see rows 4 and 9). Finally, rows 5 and 10 quantify results for the use of all proposed components, i.e. domain-specific super-resolution described in~Sec.~\ref{sec:sr} feeding into altitude-augmented detection. This approach can demonstrably and consistently outperform the other techniques across both datasets.}\vspace{25pt}
\label{table:results}
\end{table*}

\textbf{Evaluation Metrics.} Available state-of-the-art baselines for the SAVMAP and AED differ with respect to the exact evaluation metrics used. In order to address this, we provide two mean average precision~(mAP) measures for evaluation, derived slightly differently and allowing for comparability with different previous works. 

First, in line with most object detection evaluation paradigms~\cite{Redmon_2018_YOLOv3}, we provide mAP detection success judged based on the intersection-over-union~(IoU)~\cite{Redmon_2018_YOLOv3} between ground truth and candidate detection marked as mAP(IoU). In essence, a detection is a true positive only if the IoU to a ground truth annotation is above a threshold, which was set to 0.3 in this paper. If the predicted bounding box does not have a high enough IoU with any ground truth bounding box, it is classified as a false positive. Additionally, if there is no detection with a high enough IoU for a ground truth bounding box, that is a false negative. 

On the other hand, AED~\cite{Naude_2019_AED} has so far been benchmarked based on Chebyshev distances rather than IoU calculations. The Chebyshev distance can be computed as~$d = max(|x1 - x2|, |y1 - y2|)$, where $(x1, y1)$ and $(x2, y2)$ are the coordinates of the center points of the detection and ground truth, respectively. Following~\cite{Naude_2019_AED}, the detection threshold is set to 200 pixels maximally accepted distance and the metric is marked as mAP(Che) in this paper. Note that the scale of detections is ignored in this metric.

\begin{table*}[ht]
\begin{center}
\begin{tabular}{|c|l|c|c|c|c|}
\hline
\textbf{Dataset}                  & \textbf{Method}              & \textbf{Operational Resolution}                          & \textbf{Scale}                     & \textbf{mAP}(IoU)            & \textbf{mAP}(Che)       \\ \hline\hline
\multirow{12}{*}{\small{SAVMAP \cite{Reinhard_2015_SAVMAP}}}          & \cellcolor{gray!60}Baseline & \cellcolor{gray!60}                           & \cellcolor{gray!60} & \cellcolor{gray!60}0.612                & \cellcolor{gray!60}0.795                \\ \cline{2-2} \cline{5-6} 
                         & \cellcolor{gray!60} + Altitude-augmented           &\small{\multirow{-2}{*}{\cellcolor{gray!60}{$256\times 256$}}}                            & \cellcolor{gray!60}{}                          &\cellcolor{gray!60} 0.653                &\cellcolor{gray!60} 0.824                \\ \cline{2-3} \cline{5-6} 
                         & \cellcolor{gray!60} + Bicubic + Altitude-augmented   & \cellcolor{gray!60}   &\cellcolor{gray!60}                           &\cellcolor{gray!60} 0.661                &\cellcolor{gray!60} 0.834                \\ \cline{2-2} \cline{5-6} 
                         & \cellcolor{gray!60} \textbf{Ours (all components)}       &\small{\multirow{-2}{*}{\cellcolor{gray!60}$256\times 256 \rightarrow 512\times 512$}}    &\multirow{-4}{*}{\cellcolor{gray!60}{1/2 to 1}}                           &\cellcolor{gray!60} {\textbf{0.670}}  & \cellcolor{gray!60}{\textbf{0.839}} \\ \cline{2-6} 
                         & \cellcolor{gray!40} Baseline                & \cellcolor{gray!40}                             & \cellcolor{gray!40}  & \cellcolor{gray!40} 0.568                & \cellcolor{gray!40} 0.732                \\ \cline{2-2} \cline{5-6} 
                         & \cellcolor{gray!40} + Altitude-augmented           & \small{\multirow{-2}{*}{\cellcolor{gray!40} $128\times 128$}}                           & \cellcolor{gray!40}                          & \cellcolor{gray!40} 0.603                & \cellcolor{gray!40} 0.755                \\ \cline{2-3} \cline{5-6} 
                         & \cellcolor{gray!40} + Bicubic + Altitude-augmented   & \cellcolor{gray!40}    & \cellcolor{gray!40}                           & \cellcolor{gray!40} 0.625                & \cellcolor{gray!40} 0.759                \\ \cline{2-2} \cline{5-6} 
                         & \cellcolor{gray!40} \textbf{Ours (all components)}       & \small{\multirow{-2}{*}{\cellcolor{gray!40} $128\times 128 \rightarrow 512\times 512$}}   & \multirow{-4}{*}{\cellcolor{gray!40} 1/4 to 1}                          &\cellcolor{gray!40} {\textbf{0.642}} & \cellcolor{gray!40}{\textbf{0.768}} \\ \cline{2-6} 
                         & \cellcolor{gray!20} Baseline                & \cellcolor{gray!20}                             & \cellcolor{gray!20} & \cellcolor{gray!20}0.552                & \cellcolor{gray!20}0.695                \\ \cline{2-2} \cline{5-6} 
                         & \cellcolor{gray!20}+ Altitude-augmented           & \small{\multirow{-2}{*}{\cellcolor{gray!20}$64\times 64$}}                            & \cellcolor{gray!20}                          & \cellcolor{gray!20}0.594                & \cellcolor{gray!20}0.734                \\ \cline{2-3} \cline{5-6} 
                         & \cellcolor{gray!20}+ Bicubic + Altitude-augmented   &\cellcolor{gray!20}      &  \cellcolor{gray!20}                         & \cellcolor{gray!20}0.599                & \cellcolor{gray!20}0.755                \\ \cline{2-2} \cline{5-6} 
                         & \cellcolor{gray!20}\textbf{Ours (all components)}       & \small{\multirow{-2}{*}{\cellcolor{gray!20}$64\times 64 \rightarrow 512\times 512$}}     & \multirow{-4}{*}{\cellcolor{gray!20}1/8 to 1}                          & \cellcolor{gray!20}{ \textbf{0.615}} & \cellcolor{gray!20}{\textbf{0.762}} \\ \hline\hline
\multirow{12}{*}{\small{AED \cite{Naude_2019_AED}}}    & \cellcolor{gray!60}Baseline                & \cellcolor{gray!60}                           &\cellcolor{gray!60}  &\cellcolor{gray!60} 0.652                & \cellcolor{gray!60}0.872                \\ \cline{2-2} \cline{5-6} 
                         & \cellcolor{gray!60}+ Altitude-augmented           & \small{\multirow{-2}{*}{\cellcolor{gray!60}$256\times 256$}}                           &\cellcolor{gray!60}                           & \cellcolor{gray!60}0.688                & \cellcolor{gray!60}0.901                \\ \cline{2-3} \cline{5-6} 
                         & \cellcolor{gray!60}+ Bicubic + Altitude-augmented   &\cellcolor{gray!60}    & \cellcolor{gray!60}                          & \cellcolor{gray!60}0.698                & \cellcolor{gray!60}0.911                \\ \cline{2-2} \cline{5-6} 
                         & \cellcolor{gray!60}\textbf{Ours (all components)}       & \small{\multirow{-2}{*}{\cellcolor{gray!60}$256\times 256 \rightarrow 512\times 512$}}   & \multirow{-4}{*}{\cellcolor{gray!60}1/2 to 1}                          &\cellcolor{gray!60} {\textbf{0.703}} & \cellcolor{gray!60}{\textbf{0.915}} \\ \cline{2-6} 
                         & \cellcolor{gray!40}Baseline                & \cellcolor{gray!40}                           &\cellcolor{gray!40}  & \cellcolor{gray!40}0.435                & \cellcolor{gray!40}0.662                \\ \cline{2-2} \cline{5-6} 
                         & \cellcolor{gray!40}+ Altitude-augmented           & \small{\multirow{-2}{*}{\cellcolor{gray!40}$128\times 128$}}                           & \cellcolor{gray!40}                          & \cellcolor{gray!40}0.485                & \cellcolor{gray!40}0.695                \\ \cline{2-3} \cline{5-6} 
                         & \cellcolor{gray!40}+ Bicubic + Altitude-augmented   & \cellcolor{gray!40}   & \cellcolor{gray!40}                          & \cellcolor{gray!40}0.495                & \cellcolor{gray!40}0.701                \\ \cline{2-2} \cline{5-6} 
                         & \cellcolor{gray!40}\textbf{Ours (all components)}       & \small{\multirow{-2}{*}{\cellcolor{gray!40}$128\times 128 \rightarrow 512\times 512$}}   & \multirow{-4}{*}{\cellcolor{gray!40}1/4 to 1}                          & \cellcolor{gray!40}{\textbf{0.532}} & \cellcolor{gray!40}{\textbf{0.712}} \\ \cline{2-6} 
                         & \cellcolor{gray!20}Baseline                & \cellcolor{gray!20}                             & \cellcolor{gray!20} & \cellcolor{gray!20}0.312                & \cellcolor{gray!20}0.534                \\ \cline{2-2} \cline{5-6} 
                         & \cellcolor{gray!20}+ Altitude-augmented           & \small{\multirow{-2}{*}{\cellcolor{gray!20}$64\times 64$}}                             & \cellcolor{gray!20}                          & \cellcolor{gray!20}0.384                & \cellcolor{gray!20}0.585                \\ \cline{2-3} \cline{5-6} 
                         & \cellcolor{gray!20}+ Bicubic + Altitude-augmented   & \cellcolor{gray!20}     &  \cellcolor{gray!20}                         & \cellcolor{gray!20}0.452                & \cellcolor{gray!20}0.612                \\ \cline{2-2} \cline{5-6} 
                         & \cellcolor{gray!20}\textbf{Ours (all components)}       & \small{\multirow{-2}{*}{\cellcolor{gray!20}$64\times 64 \rightarrow 512\times 512$}}     & \multirow{-4}{*}{\cellcolor{gray!20}1/8 to 1}                          & \cellcolor{gray!20}{\textbf{0.475}} & \cellcolor{gray!20}{\textbf{0.633}} \\ \hline
\end{tabular}
\end{center}
\caption{\textbf{Resolution Analysis.} Artificially downsampling $512\times 512$ pixel test images via bicubic interpolation systematically simulates acquisition at lower and lower animal resolution. We show here that reconstruction back to $512\times 512$ pixel resolution via our suggested approach can maintain detection performance in these scenarios best. Superior results are consistent across datasets and mAP benchmarks for both IoU and Chebychev calculations.}
\label{table:analysis}
\end{table*}

\begin{figure*}[ht]
\begin{center}\vspace{-4pt}
\includegraphics[width=498pt,height=534pt]{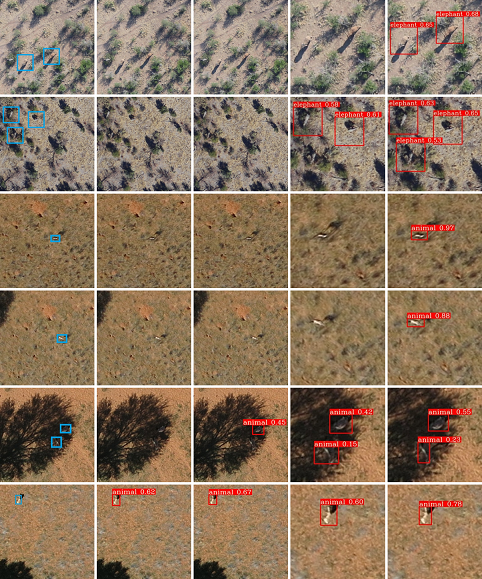}\vspace{-11pt}\
\end{center}
\textcolor{white}{-------}\textit{Ground Truth}\textcolor{white}{-----------------}\textit{Baseline}
\textcolor{white}{-----------}\textit{ Altitude-augmented}
\textcolor{white}{-------}\textit{Bicubic + Altitude-}
\textcolor{white}{----}\textit{Ours (all components)}\\ \vspace{-5pt}
\textcolor{white}{------------------------------------------------------------------------------------}\textcolor{white}{-------------.-}\textit{augmented}
\vspace{12pt}
   \caption{\textbf{Detection Examples across Methods.} Ground truth annotations of animals~(blue) in test patches from the AED (rows 1-2) and SAVMAP (rows 3-6) datasets are shown in the leftmost column. Detections~(red) and associated confidence values produced by the various methods are given in subsequent columns. The 2 rightmost columns are shown 2-fold super-resolved by associated methods in accordance with their effective operational resolution. The first 4 rows show examples where only our full approach of combined HAN super-resolution and altitude utilisation allows for correct detection. The positive effect of trivial scale-up is exemplified in row 2. Row~5 depicts a case where altitude information is critical to focus the detector on expected animal sizes; the addition of super-resolution then solves the detection problem fully. Finally, row 6 shows a common case where off-the-shelf \mm{YOLOv3 baseline} is adequate. However, note that even this case shows an improvement in detector confidence for our full approach found consistently across all examples depicted.}\vspace{-25pt}
\label{fig:visual result}
\end{figure*}

\begin{figure*}[t]
\begin{center}
\includegraphics[width=281pt]{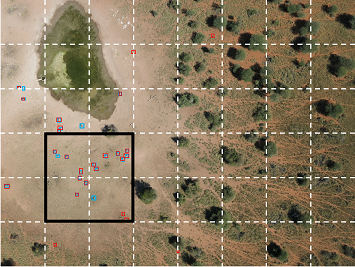}\textcolor{white}{-}\includegraphics[width=212pt]{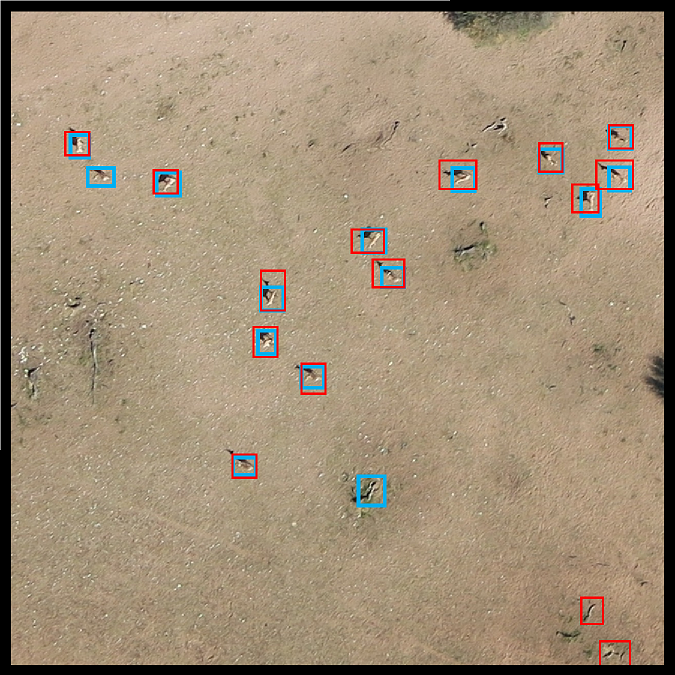}\vspace{2pt} 
\includegraphics[width=60pt]{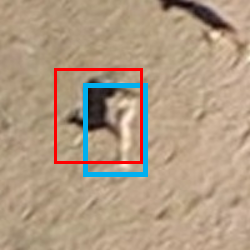}
\includegraphics[width=60pt]{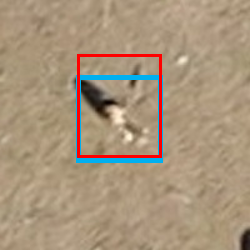}
\includegraphics[width=60pt]{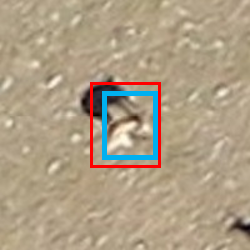}
\includegraphics[width=60pt]{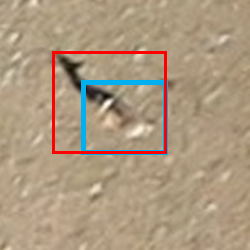}
\includegraphics[width=60pt]{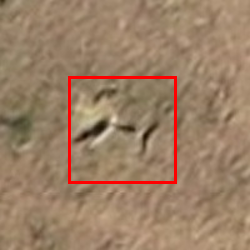}
\includegraphics[width=60pt]{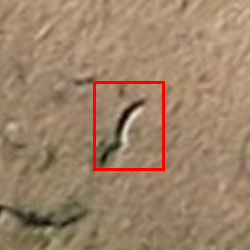}
\includegraphics[width=60pt]{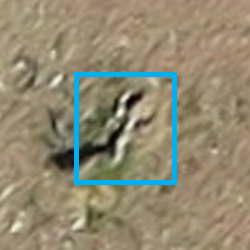}
\includegraphics[width=60pt]{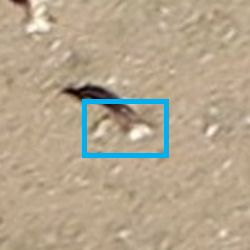}
\vspace{-21pt}
\end{center}
   \caption{\textbf{Application and Limitations - Complete SAVMAP Frame Example.} \textbf{\textit{(top left)}}~Visualisation of ground truth annotations~(blue) and detections by our proposed approach using all components~(red) on one full SAVMAP test data frame shown here in original resolution. \textbf{\textit{(top right)}}~HAN super-resolved and zoomed-in area detail with ground truth and detections covering four selected SAVMAP frame subcells. \textbf{\textit{(bottom)}}~Further zoom into some of the true positives, false positives and false negatives for best visual appreciation. Note the visual similarity of animals and other structures. Super-resolution and altitude data exploitation can only address these visual ambiguities to some extent given that even manual animal identification is extremely difficult. Therefore, additional sensor information and methodological advances are required to resolve these ambiguities further and improve survey data processing beyond the results shown in this paper.}\vspace{-5pt}
\label{fig:full}
\end{figure*}

\section{Results}

{Applying our trained framework to a single SAVMAP frame takes 3.17 seconds for the full super-resolution and detection inference process on a system with a GTX1660 Titan GPU, 16GB RAM and Intel i7-10750H CPU.}

Comparative results for both AED and SAVMAP datasets are shown in Table~\ref{table:results} whilst result independence from the choice of IoU threshold is exemplified in Fig.~\ref{fig:IoU_mAP}.  We report two different mAP outcomes underpinned by IoU and Chebychev calculations, respectively. Results show that the performance of benchmarks published so far~(rows 1 and 6) can be improved upon by utilising up-to-date YOLOv3 detectors off-the-shelf~(see rows 2 and 7), confirming efficacy arguments in~\cite{Torney_2019_comparision_DL_citizens,review2021} on our datasets.

Next, we demonstrate that the proposed use of altitude meta-data information~(see~Sec.~\ref{sec:altitude}) can consistently benefit detection performance as shown in rows 3 and 8. The addition of domain-specific super-resolution (see~Sec.~\ref{sec:sr}), as shown in rows 5 and 10 of Table~\ref{table:results}, outperforms these approaches and naive bicubic interpolation shown in rows 4 and 9, respectively. Increasing the size of the images fed into the detector from $512\times 512$ to $1024\times 1024$ increases every metric regardless of the applied method. We experimented with further upscaling, but found no significant effect. Essentially though, utilising both deep super-resolution and altitude information demonstrably and consistently outperforms other techniques across both datasets. Fig.~\ref{fig:visual result} provides qualitative examples of this superior detection performance highlighting scenarios in which the proposed concepts succeed in improving detection. Fig.~\ref{fig:full} depicts animal detection using our proposed approach on an entire SAVMAP data frame reflecting on its performance and pointing out remaining challenges. In order to investigate the efficacy of super-resolution on input image sizes further and quantify the limits of the approach, we downsampled the test portions of the SAVMAP and AED datasets across factors $\times 2$, $\times 4$ and $\times 8$ via bicubic interpolation to systematically simulate acquisition at even lower and lower animal resolution. We then reconstructed the original size via multi-scale domain-specific super-resolution. Results are presented in Table~\ref{table:analysis}. The benchmarks demonstrate that reconstruction via our suggested approach can maintain detection performance in low resolution scenarios best across all settings tested.

\section{Conclusion}
Aerial animal surveillance is an essential tool to study biodiversity and protect animal populations -- which constitutes an ethical imperative. Here, we addressed the problem of tiny animal resolutions for the first time explicitly: we combined HAN super-resolution with altitude data exploitation and showed that the integration of these components into standard recognition pipelines can systematically increase the  detection  efficacy on real-world animal datasets. {We conclude that the techniques investigated are useful tools for aerial census and conservation automation.}


{\small
\bibliographystyle{ieee_fullname}
\bibliography{bib}
}

\end{document}